\documentclass[]{spie}  
\usepackage[colorlinks,linkcolor=blue]{hyperref}
\usepackage{setspace}
 
\usepackage{amsmath,amsfonts,amssymb}
\usepackage{graphicx}
\usepackage{multirow}
\usepackage[T1]{fontenc}
\usepackage{booktabs}
\usepackage{hyperref}
\usepackage{subfigure}
\usepackage{subcaption}
\usepackage{float}

\hypersetup{hidelinks,
	colorlinks=true,
	allcolors=blue,
	pdfstartview=Fit,
	breaklinks=true}

\usepackage{setspace}

\title{Ambient Denoising Diffusion Generative Adversarial Networks for Establishing Stochastic Object Models from Noisy Image Data}

\author[a]{Xichen Xu}
\author[b]{Wentao Chen}
\author[c,d]{Weimin Zhou}
\affil[a]{Global Institute of Future Technology, Shanghai Jiao Tong University, Shanghai, 200240, China}
\affil[b]{University of Michigan-Shanghai Jiao Tong University Joint Institute, Shanghai Jiao Tong University, Shanghai, 200240, China}
\affil[c]{Department of Medical Imaging, University of Arizona, Tucson, AZ 85721, USA}
\affil[d]{Wyant College of Optical Sciences, University of Arizona, Tucson, AZ 85721, USA}

\authorinfo{Further author information: (Send correspondence to Weimin Zhou.)\\Weimin Zhou: E-mail: weiminzhou@arizona.edu}

\begin{document} 
	\maketitle
	
\begin{abstract}
It is widely accepted that medical imaging systems should be objectively assessed via task-based image quality (IQ) measures that ideally account for all sources of randomness in the measured image data, including the variation in the ensemble of objects to be imaged. Stochastic object models (SOMs) that can randomly draw samples from the object distribution can be employed to characterize object variability. To establish realistic SOMs for task-based IQ analysis, it is desirable to employ experimental image data. However, experimental image data acquired from medical imaging systems are subject to measurement noise. Previous work investigated the ability of deep generative models (DGMs) that employ an augmented generative adversarial network (GAN), AmbientGAN, for establishing SOMs from noisy measured image data. Recently, denoising diffusion models (DDMs) have emerged as a leading DGM for image synthesis and can produce superior image quality than GANs. However, original DDMs possess a slow image-generation process because of the Gaussian assumption in the denoising steps. More recently, denoising diffusion GAN (DDGAN) was proposed to permit fast image generation while maintain high generated image quality that is comparable to the original DDMs. In this work, we propose an augmented DDGAN architecture, Ambient DDGAN (ADDGAN), for learning SOMs from noisy image data. Numerical studies that consider clinical computed tomography (CT) images and digital breast tomosynthesis (DBT) images are conducted. The ability of the proposed ADDGAN to learn realistic SOMs from noisy image data is demonstrated. It has been shown that the ADDGAN significantly outperforms the advanced AmbientGAN models for synthesizing high resolution medical images with complex textures.
\end{abstract}
	
\keywords{Stochastic object model, deep generative model, denoising diffusion model, generative adversarial network}

\section{Purpose}
The objective assessment and optimization of medical imaging systems should be guided by task-based image quality (IQ) metrics, which quantify the observer performance in a clinically relevant task \cite{barrett2013foundations}. To compute task-based IQ metrics, all sources of variability in the measured data should ideally be accounted for, including the variation in the ensemble of to-be-imaged objects. To characterize object variability, stochastic object models (SOMs) that can produce a group of objects with prescribed statistical properties can be established. A well-established SOM can permit the computation of ideal observer for task-based IQ assessment \cite{zhou2019approximating, zhou2020approximating, zhou2023ideal}. Mathematical SOMs have been designed for observer studies, such as a binary texture model \cite{abbey2008ideal} and a lumpy object model \cite{rolland1992effect}. However, to capture more realistic variations of textures and anatomical structures, it is desirable to establish SOMs based on experimental data. Kupinski \emph{et al.} \cite{kupinski2003experimental} proposed a method to ﬁt object models into noisy image data. However, this method requires that the characteristic functional of random objects can be analytically determined.

Deep generative models (DGMs), such as generative adversarial networks (GANs), have emerged as a powerful tool for learning image distributions from training data. To establish SOMs from noisy measured image data, AmbientGANs \cite{bora2018ambientgan,zhou2022learning,xu2024ambientcyclegan,chen2024ambient} were proposed. More recently, a novel paradigm of DGM, denoising diffusion models (DDMs), shows superior image synthesis performance compared with GANs \cite{deshpande2024assessing}. However, the original DDMs require numerous denoising steps to satisfy a Gaussian assumption, leading to a long-sampling process. To accelerate the sampling process, denoising diffusion GAN (DDGAN) \cite{xiao2021tackling} was proposed, which significantly reduces the number of denoising steps by utilizing a GAN that models a complex multi-modal denoising distribution. However, to establish SOMs, DDGANs cannot be directly trained on medical image data that are contaminated by noise.

In this paper, we propose a novel augmented DDGAN architecture, Ambient Denoising Diffusion GAN (ADDGAN), to establish realistic SOMs from noisy medical image data. Preliminary studies that considered a clinic computed tomography (CT) dataset and digital breast tomosynthesis (DBT) dataset were conducted. It is demonstrated that the proposed ADDGAN can successfully produce superior image quality than AmbientGAN.

\section{Methods}
\subsection{GANs and DDMs}
\noindent\textbf{Generative adversarial network (GAN)} is an implicit DGM to learn the image distribution and can sample new images. A traditional GAN consists of two neural networks, a generator \(G\) and a discriminator \(D\). The \(G\) generates synthesized images by taking a latent vector as input, and \(D\) tries to distinguish between real and synthesized images. The adversarial training process can be described as:

\begin{equation}
\min_{G} \max_{D} \mathbb{E}_{\mathbf{x} \sim p(\mathbf{x})} [\log D(\mathbf{x})] + \mathbb{E}_{\mathbf{z} \sim p(\mathbf{z})} [\log (1 - D(G(\mathbf{z})))],
\label{eq_gen}
\end{equation}

\noindent where \(\mathbf{x}\) represents real images from the distribution \(p(\mathbf{x})\), and \(\mathbf{z}\) represents latent vectors sampled from a prior distribution \(p{(\mathbf{z})}\), such as Gaussian distribution.

\noindent\textbf{Denoising diffusion models (DDMs)} have emerged as leading DGMs in image synthesis. DDMs construct a unique mapping between the isotropic Gaussian noise and real images by progressively denoising over a sequence of discrete time steps $\{0,1,\ldots,t\}$. A conventional DDM contains a forward process and a reverse process. In the forward process, Gaussian noise is gradually added to the data, while the reverse process aims to denoise and reconstruct the data. The two processes can be represented as:

\begin{equation}
\begin{aligned}
\text{Forward process: }q(\mathbf{x}_t \mid \mathbf{x}_{t-1}) &= \mathcal{N}(\mathbf{x}_t; \sqrt{1 - \beta_t} \mathbf{x}_{t-1}, \beta_t \mathbf{I}),\\
\text{Reverse process: }p_{\theta}(\mathbf{x}_{t-1} \mid \mathbf{x}_t) &= \mathcal{N}(\mathbf{x}_{t-1}; \mu_{\theta}(\mathbf{x}_t, t), \Sigma_{\theta}(\mathbf{x}_t, t)),
\label{eq_ddpm}
\end{aligned}
\end{equation}

\noindent where \(\mathbf{x}_t\) and \(\mathbf{x}_{t-1}\) are noisy images at the $t$-th step and the $(t-1)$-th step, \(\beta_t\) is a variance schedule, \(p_{\theta}\) is the learned denoising Gaussian model that has a learnable mean \(\mu_{\theta}\) and covariance matrix \(\Sigma_{\theta}\). Here $\theta$ represents the parameters of the neural networks. More details about DDMs can be found in the paper \cite{ho2020denoising}. Although DDM can outperform GAN in image synthesis, its image-generation process is typically very slow.

\subsection{DDGAN}
\noindent\textbf{Denoising diffusion GAN (DDGAN)} employes a generator $G_\theta(\mathbf{x}_t,\mathbf{z},t)$ to capture the complex multimodal denoising distribution, avoiding the Gaussian assumption required between the adjacent steps in the reverse process of DDM. Meanwhile, the discriminator $D_\phi(\mathbf{x}_t,\mathbf{x}_{t-1},t)$ determines whether the image $\mathbf{x}_{t-1}$ is a reasonable denoised approximation of $\mathbf{x}_{t}$. The $D_\phi(\mathbf{x}_t,\mathbf{x}_{t-1},t)$ can be trained according to Eq. (\ref{eq3}):

\begin{equation}
\min_\phi \sum_{t \geq 1} \{\mathbb{E}_{q\left(\mathbf{x}_0\right)q\left(\mathbf{x}_{t-1}\mid\mathbf{x}_0\right)q\left(\mathbf{x}_t\mid\mathbf{x}_{t-1}\right)}[-\log(D_\phi(\mathbf{x}_t,\mathbf{x}_{t-1},t))]+\mathbb{E}_{q\left(\mathbf{x}_t\right)p_\theta\left(\mathbf{x}_{t-1}\mid\mathbf{x}_t\right)}[-\log(1-D_\phi(\mathbf{x}_t,\mathbf{x}_{t-1},t))]\}.
\label{eq3}
\end{equation}

\noindent The $G_\theta(\mathbf{x}_t,\mathbf{z},t)$ is trained according to Eq. (\ref{eq_G}):

\begin{equation}
\max _\theta \sum_{t\geq 1} \mathbb{E}_{q\left(\mathbf{x}_t\right)} \mathbb{E}_{p_\theta\left(\mathbf{x}_{t-1} \mid \mathbf{x}_t\right)}\left[\log \left(D_\phi\left(\mathbf{x}_{t}, \mathbf{x}_{t-1}, t\right)\right)\right],
\label{eq_G}
\end{equation}

\noindent where $p_\theta\left(\mathbf{x}_{t-1}\mid\mathbf{x}_t\right)=\int p\left(\mathbf{z}\right)q (\mathbf{x}_{t-1}\mid\mathbf{x}_t,\mathbf{\hat{x}}_0=G_\theta(\mathbf{x}_t,\mathbf{z},t))~d\mathbf{z}$ is a parameterized denoising model, $G_\theta(\mathbf{x}_t,\mathbf{z},t)$ and $D_\phi(\mathbf{x}_t,\mathbf{x}_{t-1},t)$ are generator and discriminator, respectively. Here, $\theta$ and $\phi$ represent the weight parameters of the networks, $\mathbf{x}_{t-1}$ and $\mathbf{x}_{t}$ are degraded images in the $t$-th and $(t-1)$-th time steps in the forward process, $\mathbf{x}_{0}$ is considered as the clean image, and $\mathbf{z}$ follows a standard Gaussian distribution, i.e., $\mathbf{z}\sim\mathcal{N}\left(\mathbf{0},\mathbf{I}\right)$. However, DDGAN cannot be directly trained on noisy image data to establish SOMs that should be independent on measurement noise.

\subsection{Ambient Denoising Diffusion GAN}	
\begin{figure}[htb]
		\centering
		\setlength{\abovecaptionskip}{0mm} 
		\includegraphics[width=\linewidth]{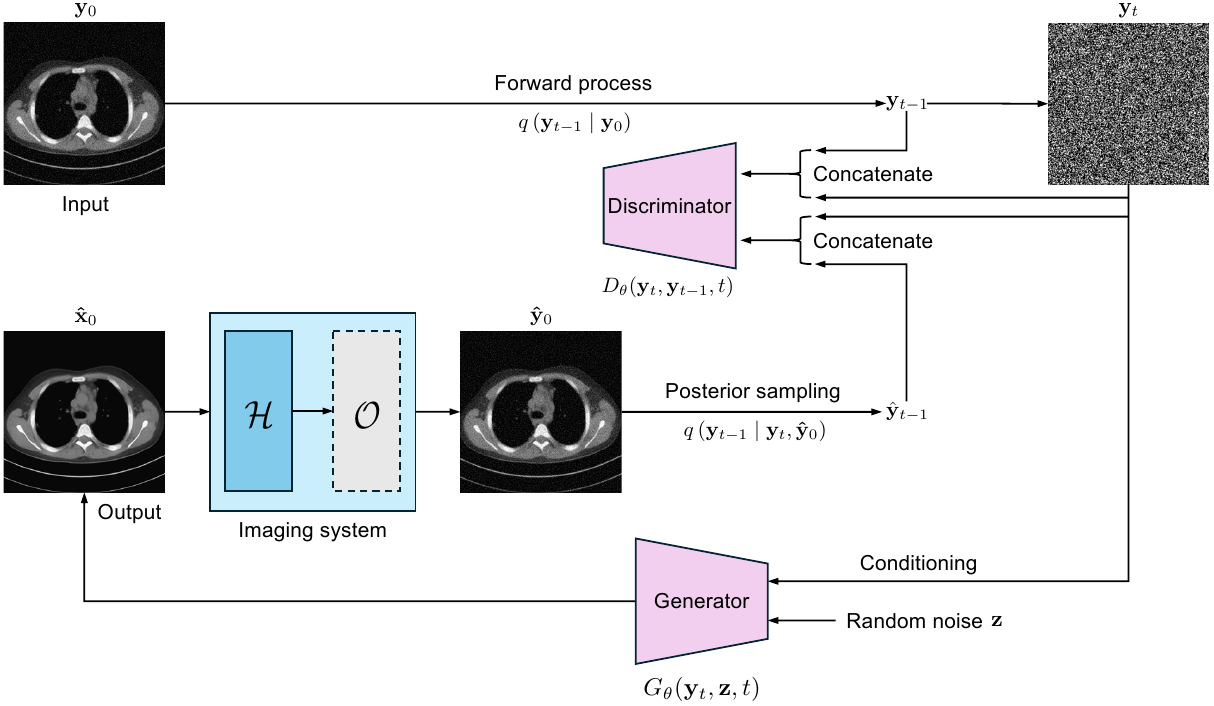}\\
		\caption{The illustration of the proposed ADDGAN architecture. In our CT case, $\mathcal{H}$ corresponds to the Radon transform, and $\mathcal{O}$ denotes the reconstruction operator.}
		\label{net}
\end{figure}

In this work, we propose an augmented DDGAN structure, Ambient Denoising Diffusion GAN (ADDGAN), which can be trained directly on noisy medical image data for establishing realistic SOMs. The architecture of the ADDGAN is illustrated in Fig. \ref{net}.

The noisy measured data and the corresponding to-be-imaged object are denoted as $\mathbf{y}_{0}$ and $\mathbf{x}_{0}$, respectively. When computed imaging systems are considered, $\mathbf{y}_{0}$ is the reconstructed image. The generator $G_\theta(\mathbf{y}_{t},\mathbf{z},t)$ in the proposed ADDGAN parametrizes the denoising distribution $p_\theta\left(\mathbf{y}_{t-1}\mid\mathbf{y}_t\right)$ for each reverse step. Specifically, $G_\theta(\mathbf{y}_{t},\mathbf{z},t)$ is responsible for mapping the degraded measured data $\mathbf{y}_{t}$, which are generated in the forward process, to the corresponding synthesized objects $\hat{\mathbf{x}}_{0}$. Subsequently, the pre-known imaging operator $\mathcal{H}$ (and the reconstruction operator $\mathcal{O}$ if computed imaging systems) is applied to $\hat{\mathbf{x}}_{0}$ to compute the simulated measured data $\hat{\mathbf{y}}_{0}$ for use in the posterior sampling to draw $\hat{\mathbf{y}}_{t-1}$. The denoising distribution $p_\theta\left(\mathbf{y}_{t-1}\mid\mathbf{y}_t\right)$ can be subsequently re-formulated as follows:
\small{
\begin{equation}
p_\theta\left(\mathbf{y}_{t-1}\mid\mathbf{y}_t\right)=\int p\left(\mathbf{z}\right)q\left(\mathbf{y}_{t-1}\mid\mathbf{y}_t,\mathbf{\hat{y}}_0\right)~d\mathbf{z}=\int p\left(\mathbf{z}\right)q\left(\mathbf{y}_{t-1}\mid\mathbf{y}_t,\mathcal{H} \left(\mathbf{\hat{x}}_0\right)\right)~d\mathbf{z} = \int p\left(\mathbf{z}\right)q\left(\mathbf{y}_{t-1}\mid\mathbf{y}_t,\mathcal{H} \left(G_\theta(\mathbf{x}_{t},\mathbf{z},t)\right)\right)~d\mathbf{z},
\label{addgan}
\end{equation}}

\normalsize
\noindent where $\mathbf{y}_t$ and $\mathbf{y}_{t-1}$ are the samples with $t$-step and $(t-1)$-step scheduled noise added to the input $\mathbf{y}_0$, $\mathbf{z}$ follows a Gaussian distribution, $\hat{\mathbf{x}}_{0}$ is the synthesized object, and $\hat{\mathbf{y}}_{0}$ denotes the corresponding simulated measurements. After multiple denoising steps according to $p_\theta\left(\mathbf{y}_{t-1}\mid\mathbf{y}_{t} \right)$, 
the generator-produced image $\hat{\mathbf{x}}_{0}$ that models $p_\theta(\mathbf{y}_{0}\mid\mathbf{y}_{1} )$ according to Eq. (\ref{addgan}) will be considered as the final output.

\section{Numerical studies and results}
\subsection{Datasets and implementation details}
Numerical studies that considered two different clinic datasets were conducted. In the first study, \textbf{a stylized computed tomographic imaging system} was considered. This imaging system was described as: 
\begin{equation}
    \mathbf{g}=\mathcal{R}\mathbf{f}+\mathbf{n},
\end{equation}
where $\mathcal{R}$ denotes a 2D discrete Radon transform \cite{kak2002books} associated with parallel beams that maps a 2D object $\mathbf{f}$ to a sinogram $\mathbf{g}$. The angular scanning range was 180 degrees, with tomographic views evenly spaced at a 1 degree angular step. A clinical CT dataset that contains over 22,000 clinical CT images from the DeepLesion dataset \cite{yan2018deeplesion} was employed. Those CT images were resized to the dimension of 256 $\times$ 256 to serve as a group of ground truth objects $\mathbf{f}$ and were normalized to the range between 0 and 1. A collection of measured data $\mathbf{g}$ were simulated by applying $\mathcal{R}$ on each object and adding independent and identically distributed (i.i.d.) Gaussian noise with a standard deviation of 1. From the measured data $\mathbf{g}$, a set of reconstructed objects $\mathbf{f}_{recon}$ was generated using a filtered back-projection (FBP) reconstruction algorithm that employed a Ram-Lak filter. The $\mathbf{f}_{recon}$ was used as $\mathbf{y}_{0}$ in the ADDGAN training. In the ADDGAN training process, the Radon transform $\mathcal{R}$ and the FBP operator were applied to the generated objects $\hat{\mathbf{x}}_{0}$ to produce $\hat{\mathbf{y}}_{0}$. The implementation of the CT imaging system utilized the $\verb|TorchRadon|$ library \cite{ronchetti2020torchradon}.

In the second study, digital breast tomosynthesis (DBT) images from the \textbf{BCS-DBT dataset} \cite{buda2020detection} were employed, and 8,325 DBT images were selected and resized to the dimension of 1024 $\times$ 1024. 
These images were employed to serve as a group of ground-truth objects.
In this preliminary study of DBT, we considered a simplified toy problem and directly added Gaussian noise with a mean of 0 and standard deviation (std) of 0.06 to the prepared DBT images to generate noisy image data $\mathbf{y}_{0}$.

The implementation of the proposed ADDGAN is based on the official DDGAN code\footnote{\href{https://github.com/NVlabs/denoising-diffusion-gan}{https://github.com/NVlabs/denoising-diffusion-gan}}. For CT images, the ADDGAN was trained on 4 NVIDIA GeForce RTX 3090 Ti GPUs with the batch size 16. Both the generator and discriminator were optimized using the Adam optimizer \cite{kingma2014adam} with a learning rate of 0.0002, and the diffusion process had four time steps. For DBT images, 8 GPUs were deployed for training the ADDGAN and the batch size was set to eight. The learning rates of the Adam optimizers were $1.6\times 10^{-5}$ for the generator and $1\times 10^{-5}$ for the discriminator, and the diffusion process comprised eight time steps.

\subsection{Results}
\begin{figure}[H]
		\centering
		\setlength{\abovecaptionskip}{0mm} 
		\includegraphics[width=\linewidth]{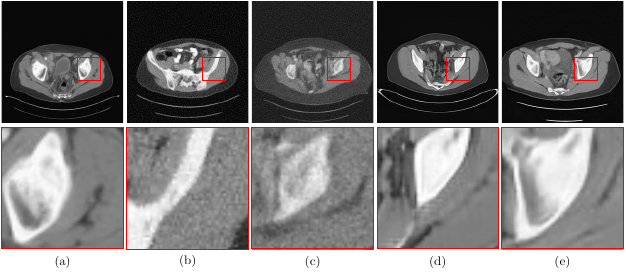}\\
		\caption{The first row shows full CT images, while the second row shows detailed texture in the red-box region. From left to right are (a) ground-truth images, (b) Denoisng Diffusion GAN-produced images, (c) DDPM-produced images, (d) ProAmGAN-produced images, (e) Proposed ADDGAN-produced images.}
		\label{result}
\end{figure}

\begin{figure}[H]
		\centering
		\setlength{\abovecaptionskip}{0mm} 
		\includegraphics[width=\linewidth]{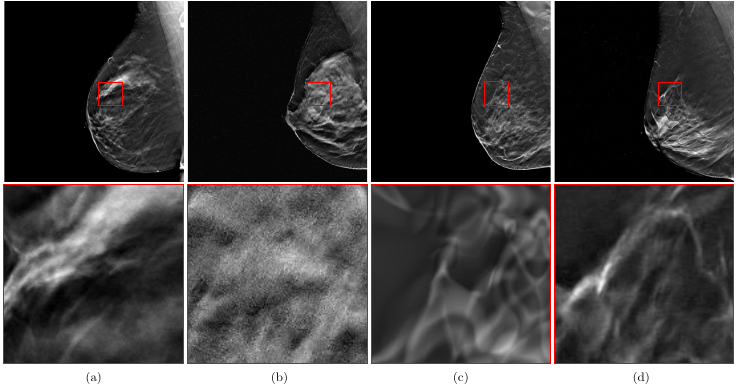}\\
		\caption{The first row shows full DBT images, while the second row are detailed texture in the red-box region. From left to right are (a) ground-truth images, (b) Denoisng-Diffusion-GAN-produced images, (c) Ambient StyleGAN3-produced images, (d) Proposed ADDGAN-produced images.}
		\label{result2}
\end{figure}

The ADDGAN was compared to the Denoising Diffusion Probabilistic Model (DDPM) \cite{ho2020denoising} and the DDGAN that were directly trained on noisy image data. In addition to these two methods, we also compared ADDGAN to two state-of-the-art AmbientGAN-based methods: Ambient StyleGAN3 \cite{zhou2022learning}, which leverages the StyleGAN3 architecture \cite{karras2021alias}, and progressive growing AmbientGAN (ProAmGAN) \cite{zhou2022learning}. Fig. \ref{result} and \ref{result2} present examples of the full objects and zoomed-in regions produced by different models for CT and DBT, respectively. The generated images from DDPM and DDGAN are strongly affected by measurement noise. Although Ambient StyleGAN3 and ProAmGAN produced clean images, the generated image textures are significantly distorted. 

\begin{figure}[H]
    \centering
    \setlength{\abovecaptionskip}{0mm} 
    \includegraphics[width=\linewidth]{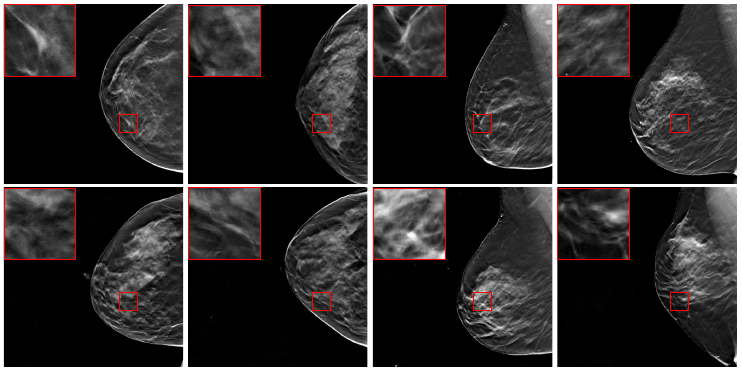}\\
    \caption{Ground truth objects (the first row) and ADDGAN-generated objects (the second row).}
    \label{detail_dbt}
\end{figure}

Examples of ADDGAN-produced DBT images are compared to the ground-truth DBT images in Fig. \ref{detail_dbt}. ADDGAN-produced DBT images have similar visual appearance with the ground-truth DBT images and possess realistic textures.

\begin{figure}[H]
\centering
\subfigure[\label{eva_ssimpdf}]{
    \begin{minipage}[t]{0.55\linewidth}
 	\centering
 	\includegraphics[width=\textwidth]{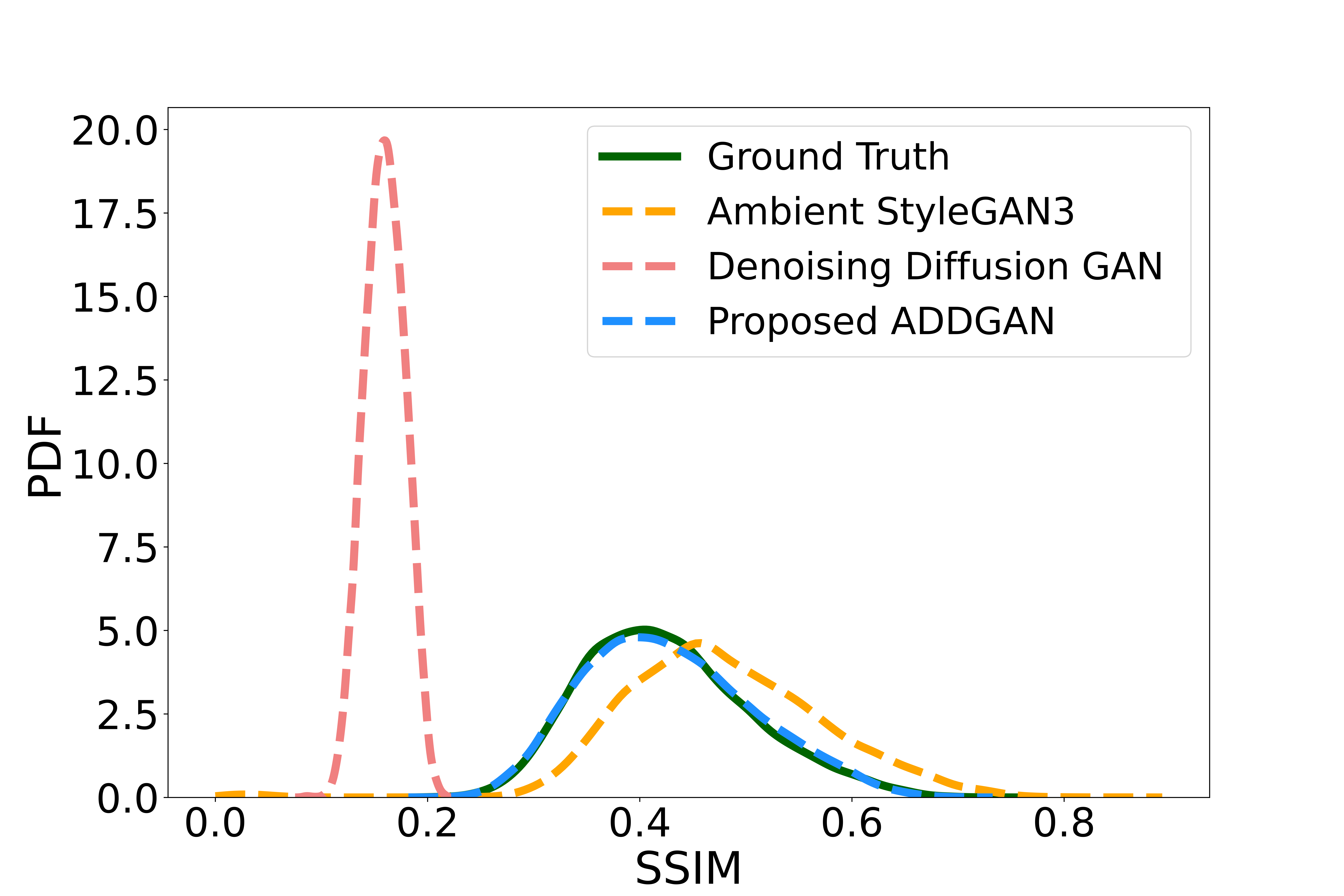}\\
 	\end{minipage}
}
\subfigure[\label{eva_roc}]{
    \begin{minipage}[t]{0.4\linewidth}
 	\centering
 	\includegraphics[width=\textwidth]{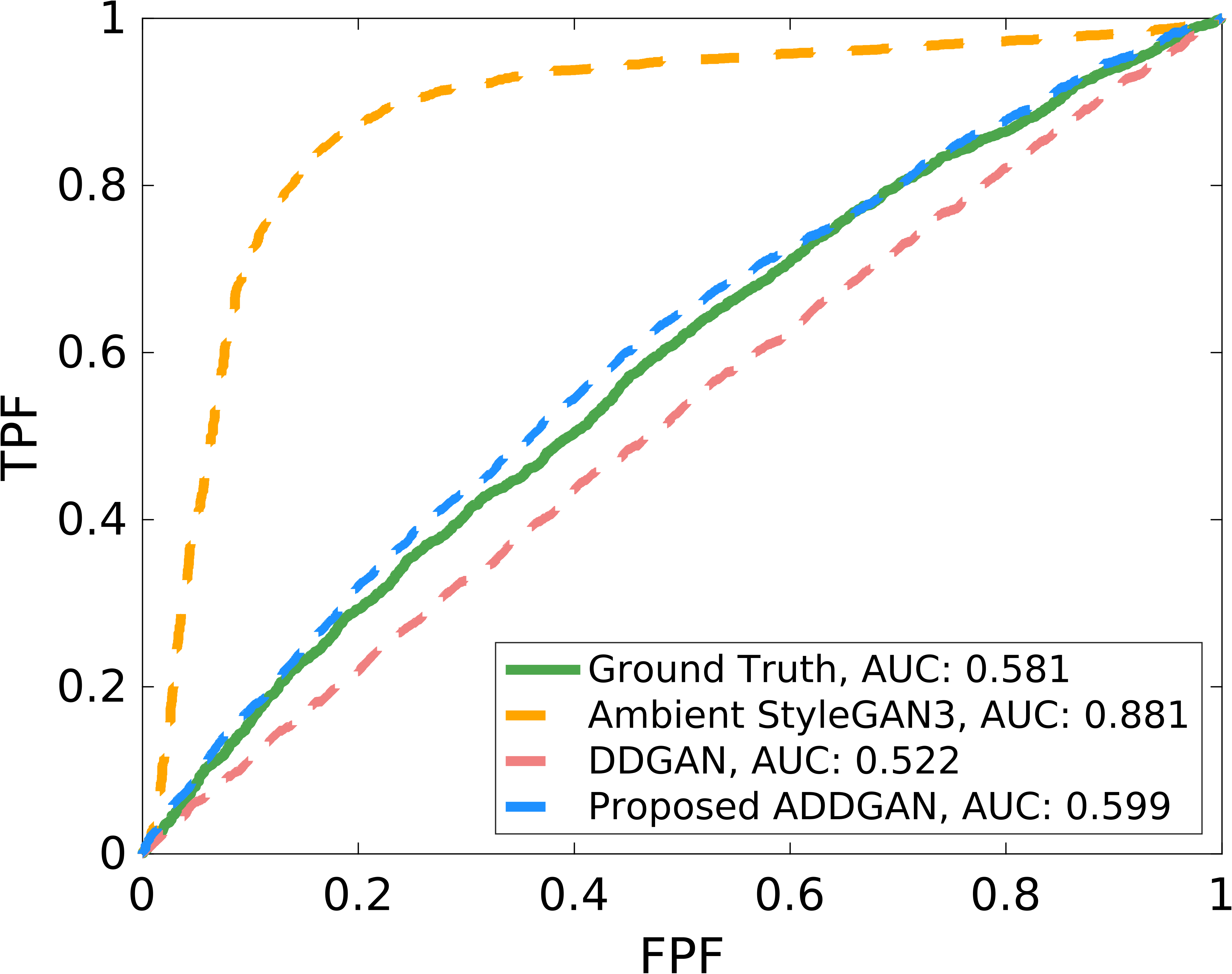}\\
 	\end{minipage}
}
\centering
\caption{(a) PDFs of SSIMs, (b) Signal-detection performance using Hotelling observer calculated on 16,000 DBT patches.}
\label{eva}
\end{figure}

The Fr{\'e}chet Inception Distance (FID) score was calculated to assess the performance of different generative models based on a set of 9,624 real images and 10,000 generated images for stylized CT dataset, and a set of 12,000 real images and 12,000 generated images for BCS-DBT dataset. Additionally, to assess the ability of the trained models to produce textures, we conducted evaluation studies on the cropped patches. Specifically, we cropped over 16,000 DBT patches of size 256 $\times$ 256. The Probability Density Functions (PDFs) of the Structural Similarity Index Measure (SSIM) values for random pairs of ground truth patches and model-produced patches were calculated. For evaluating the task-based IQ, a signal-known-exactly binary signal detection task and the Hotelling observer were considered. Specifically, backgrounds from ground-truth patches and generative model-produced patches were further cropped to 64 $\times$ 64 patches, and a Gaussian signal with a width of 0.22 and an amplitude of 0.50 was added to these patches to simulate signal-present patches. The Hotelling template corresponding to the ground-truth and model-produced images was computed by use of 16,000 patches, and the observer performance was assessed on a different set of 1,500 signal-present and 1,500 signal-absent patch samples. The FID scores are reported in Table \ref{fid_ct_10000} and Table \ref{fid_dbt_8000}. The SSIM PDFs and ROC curves are shown in Fig. \ref{eva_ssimpdf} and Fig. \ref{eva_roc}, respectively. The proposed ADDGAN achieves superior performance across all evaluation metrics. The ROC curves and AUC values in Fig. \ref{eva} indicate that the Hotelling observer performance corresponding to the ADDGAN-generated SOM is much more closely matched that of the ground truth images than those from other models.

\begin{table}[ht]
    \centering
    \begin{minipage}[t]{0.45\textwidth}
        \centering
        \begin{tabular*}{\textwidth}{@{\extracolsep{\fill}}l c}
            \toprule
            \multirow{1}{*}{Method}  
            & \multicolumn{1}{c}{FID ($\downarrow$)} \\ \midrule
            DDGAN & 90.03 \\
            DDPM & 96.38 \\
            ProAmGAN & 41.80 \\
            \midrule
            ADDGAN & \textbf{30.34} \\
            \bottomrule
        \end{tabular*}
        \caption{FID scores on 10,000 generated CT images and 9,624 real images.}
        \label{fid_ct_10000}
    \end{minipage}\hfill
    \begin{minipage}[t]{0.45\textwidth}
        \centering
        \begin{tabular*}{\textwidth}{@{\extracolsep{\fill}}l c}
            \toprule
            \multirow{1}{*}{Method}  
            & \multicolumn{1}{c}{FID ($\downarrow$)} \\ \midrule
            DDGAN & 75.10 \\
            Ambient StyleGAN3 & 70.04 \\
            \midrule
            ADDGAN & \textbf{41.13} \\
            \bottomrule
        \end{tabular*}
        \caption{FID scores on 12,000 generated DBT images and 12,000 real images.}
        \label{fid_dbt_8000}
    \end{minipage}    
    \label{fid_combined}
\end{table}

\section{Conclusions}
In this study, an augmented DDGAN architecture, Ambient Denoising Diffusion GAN (ADDGAN), was introduced for establishing realistic stochastic object models (SOMs) from noisy medical image data. Numerical studies that consider clinical CT and DBT images were systematically conducted. The established SOMs were assessed via both traditional and task-based image quality metrics. It was demonstrated that the proposed ADDGAN outperforms Ambient StyleGAN3 for synthesizing objects in medical imaging and holds great potential for establishing realistic SOMs for the objective assessment of image quality.

\bibliography{report} 
\bibliographystyle{spiebib} 
	
\end{document}